\documentclass[runningheads]{llncs}

\usepackage{eccv}

\usepackage{eccvabbrv}

\usepackage{graphicx}
\usepackage{booktabs}

\usepackage[accsupp]{axessibility}  

\usepackage{hyperref}

\usepackage{orcidlink}

\newcommand{\ours}{DyBDet}
\usepackage{multirow}
\usepackage{pifont}
\usepackage{xcolor}
\usepackage{colortbl}

\makeatletter
\newcommand\figcaption{\def\@captype{figure}\caption}
\newcommand\tabcaption{\def\@captype{table}\caption}
\makeatother

\begin{document}

\title{Fine-grained Dynamic Network for Generic Event Boundary Detection} 


\author{Ziwei Zheng\orcidlink{0009-0000-4896-3293} \and
Lijun He\orcidlink{0000-0002-3911-8263} \and
Le Yang\orcidlink{0000-0001-8379-4915} \and
Fan Li\thanks{Corresponding author}\orcidlink{0000-0002-7566-1634}}

\authorrunning{Z.~Zheng et al.}

\institute{Xi'an Jiaotong University \\
\email{ziwei.zheng@stu.xjtu.edu.cn, yangle15@xjtu.edu.cn} \\
\email{\{lijunhe,lifan\}@mail.xjtu.edu.cn}}
\maketitle

\begin{abstract}

 Generic event boundary detection (GEBD) aims at pinpointing event boundaries naturally perceived by humans, playing a crucial role in understanding long-form videos. Given the diverse nature of generic boundaries, spanning different video appearances, objects, and actions, this task remains challenging. Existing methods usually detect various boundaries by the same protocol, regardless of their distinctive characteristics and detection difficulties, resulting in suboptimal performance. Intuitively, a more intelligent and reasonable way is to adaptively detect boundaries by considering their special properties. In light of this, we propose a novel dynamic pipeline for generic event boundaries named DyBDet. By introducing a multi-exit network architecture, DyBDet automatically learns the subnet allocation to different video snippets, enabling fine-grained detection for various boundaries. Besides, a multi-order difference detector is also proposed to ensure generic boundaries can be effectively identified and adaptively processed. Extensive experiments on the challenging Kinetics-GEBD and TAPOS datasets demonstrate that adopting the dynamic strategy significantly benefits GEBD tasks, leading to obvious improvements in both performance and efficiency compared to the current state-of-the-art. The code is available at \url{https://github.com/Ziwei-Zheng/DyBDet}.

\end{abstract}


\section{Introduction}
\label{sec:intro}

Video comprehension has gained significant traction due to the rapid growth of online resources. According to cognitive science \cite{radvansky2011event}, humans naturally tend to parse long videos into meaningful segments. In light of this, Generic Event Boundary Detection \cite{shou2021generic} (GEBD) has been introduced to localize distinctive event boundaries, aiming to advance the field of long-form video understanding.

\begin{figure}
    \centering
    \includegraphics[width=1\linewidth]{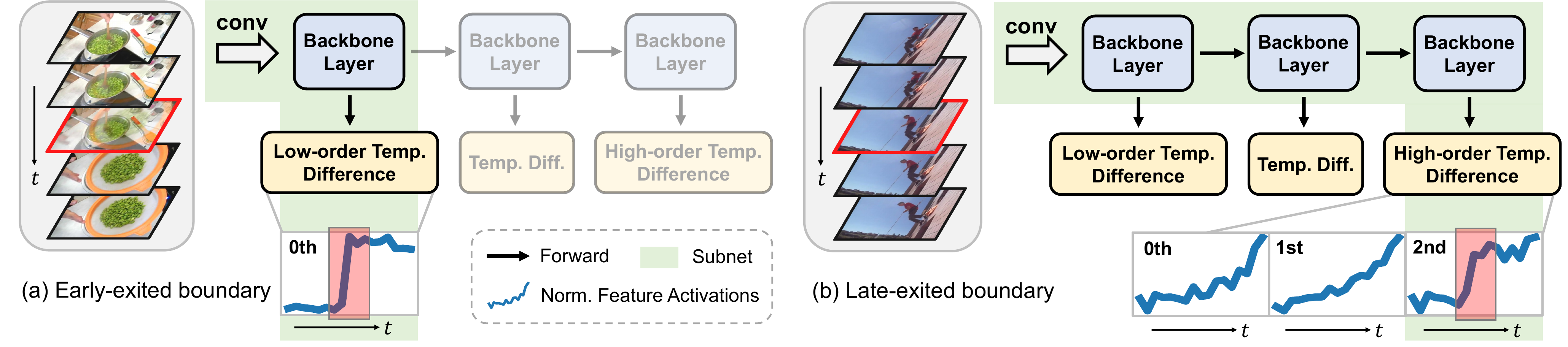}
    \caption{Adaptive inference for video snippets within the dynamic architecture. We use temporal differences with different orders to capture the distinctive boundary features and plot the normalized activations for better visualization. The ground-truth boundaries are highlighted with red lines. (a): The boundary of the shot change can be obviously identified with low-level appearance features w/o temporal difference and can exit early to save computations. (b): The action change relies on features of high-level semantics and high-order temporal differences to reveal the boundary.}
    \label{fig1}
    \vspace{-10pt}
\end{figure}

Generic event boundaries are taxonomy-free and encompass various low-level and high-level semantic changes. Modeling diverse boundaries of different semantic levels in the same way could entangle the feature extraction and result in inferior performance. For instance, the shot change in Figure \ref{fig1} (a) can be efficiently and effectively detected with appearance-level information only. In contrast, exploring high-level semantics is crucial for detecting the sub-action change in Figure \ref{fig1} (b). However, applying deep models to video snippets that contain shot changes may vanish the coarse-level but distinctive appearance features, as well as unnecessary computational waste. Therefore, detecting generic event boundaries with the same protocol regardless of their distinctive characteristics and detection difficulties can lead to suboptimal performance and efficiency.

To address this issue, we propose a dynamic pipeline capable of sample-dependent detections. A multi-exit network is proposed in \ours{}, allowing adaptive inference for various boundaries by automatically learning subnet allocation. The boundaries corresponding to low-level semantic information, such as shot changes, are only passed through the subnet with shallow layers to capture appearance information and simple temporal dependencies. While boundaries needing high-level semantics for boundary identification will traverse the whole network for in-depth spatiotemporal modeling. By inferring different inputs in a fine-grained manner, the processing of generic event boundaries can be specialized to enhance the performance and efficiency of the overall detection system.

To further ensure various boundaries can be effectively detected with specialized strategies, we propose a multi-order difference detector to explore the most distinctive localized change pattern of each candidate, which is crucial for boundary identification. Specifically, a multi-order difference encoder and a pairwise contrast module are proposed to compose the boundary detector. Detectors with different order combinations are attached at varying depths of the backbone to form the overall dynamic architecture. The temporal difference describes the ``change'' within adjacent frames, which is crucial for event boundary detection \cite{tang2022progressive}. Figure \ref{fig1} (a) shows that the shot change has a clear boundary itself based on features from shallow layers of the backbone. However, sub-action changes in Figure \ref{fig1} (b) are much more complex in time, needing extra high-order temporal differences to reveal the distinctive boundary information, which is usually overlooked by previous methods. Therefore, by introducing multi-order temporal differences to the dynamic model, various boundaries can be easily distinguished and adaptively processed, leading to substantial improvement.

To the best of our knowledge, we are the first to introduce dynamic processing for generic event boundary detection. Extensive experiments and studies on Kinetics-GEBD \cite{shou2021generic} and TAPOS \cite{shao2020intra} dataset demonstrate the effectiveness and efficiency of adaptive inference to achieve the new state-of-the-art.

\vspace{-5pt}

\section{Related Work}
\label{sec:related}

\vspace{-5pt}

\subsubsection{Temporal Detection Tasks and GEBD.} Temporal detection tasks are proposed to detect clip-level instances in untrimmed videos almost at a similar difficulty level, including shot boundary detection \cite{gygli2018ridiculously,souvcek2019transnet,tang2018fast}, temporal action segmentation \cite{aakur2019perceptual,alayrac2017joint,farha2019ms,lei2018temporal} and localization \cite{zhang2022actionformer,nag2023post,cheng2022tallformer,shi2023tridet}. \cite{shou2021generic} first proposed generic event boundary detection (GEBD) to localize the taxonomy-free moments that humans naturally perceive event boundaries. The boundaries can be used for further long-form video comprehension. Recent works either follow a similar fashion of \cite{shou2021generic} to slice the long video into adjacent overlapped snippets as independent samples \cite{tang2022progressive,hong2021generic,kang2021winning}, or take the whole video as input with continuous predictions \cite{li2022structured, huynh2023generic}. Specifically, \cite{kang2022uboco} proposes a recursive parsing algorithm based on the temporal self-similarity matrix to enhance local modeling. \cite{tang2022progressive} aims to characterize the motion pattern with dense difference maps. Unlike these methods process all boundaries with the same protocol, we propose a dynamic pipeline for specialized processing with higher performance and efficiency.

\vspace{-10pt}

\subsubsection{Dynamic Neural Networks.} Due to the favorability of efficiency while maintaining high performance, dynamic neural networks are attracting more attention. Different from the conventional methods that handle various inputs with a fixed network, dynamic networks \cite{huang2018multi,han2021dynamic,wang2018skipnet,li2020learning,zheng2023dynamic,yang2021condensenet} are capable of sample-wise adaptive inference. Each sample can receive customized processing with adapted subnets within the dynamic architecture \cite{yang2020resolution,chen2019you,wang2021adaptive,wang2022adafocus,han2024latency}.
When adopted to downstream tasks \cite{yang2023adadet,dai2021dynamic,xie2020spatially,ming2021dynamic,wang2024towards}, the subnets in dynamic networks usually still share most of the parameters for the consideration of efficiency, and the distinctive characteristics for inputs are not fully explored. Different from them, the dynamic network in our method excels in fine-grained detection, leveraging the special properties inherent in each sample.

\begin{figure}
    \centering
    \includegraphics[width=1\linewidth]{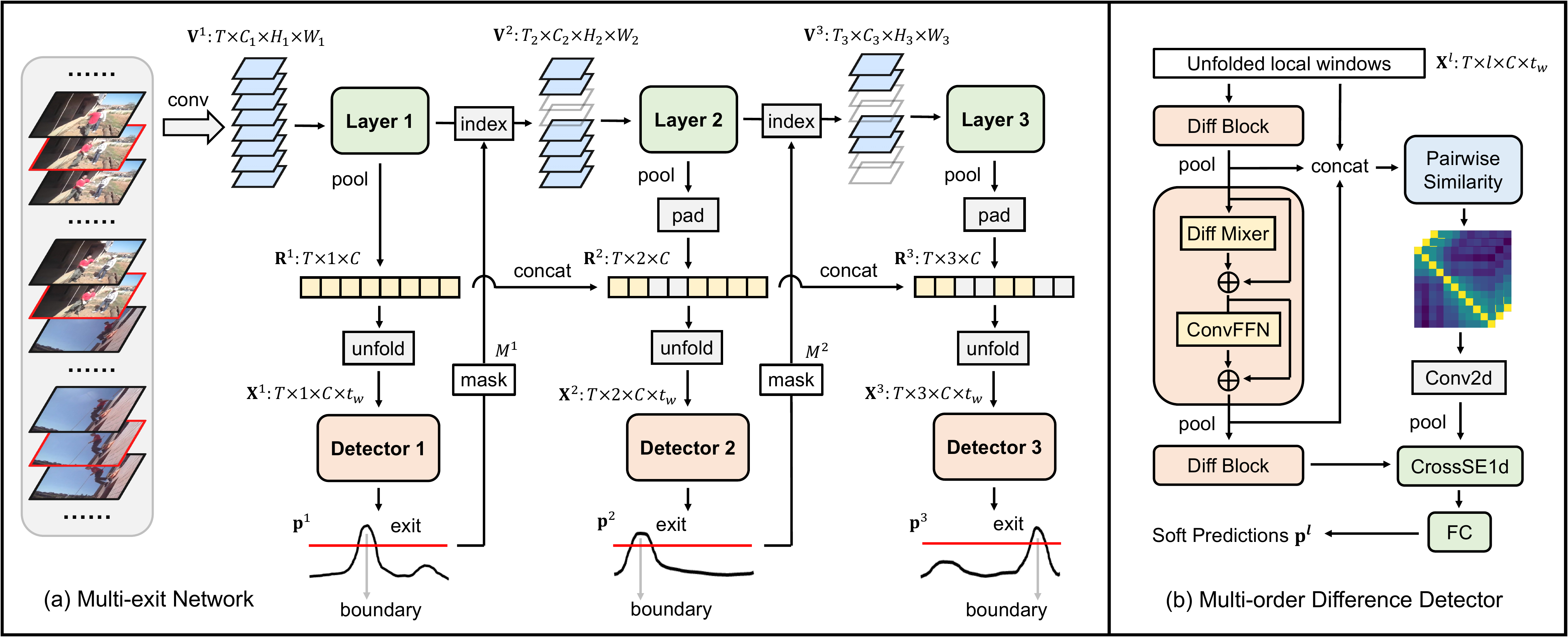}
    \caption{Overview of the proposed \ours{}. Boundaries are highlighted with red lines. (a): the \textbf{multi-exit network} to enable frame-level adaptive inference, (b): the \textbf{multi-order difference detector} to distinguish boundaries with various characteristics.}
    \label{fig_over}
    \vspace{-10pt}
\end{figure}

\vspace{-5pt}

\section{Method}
\label{sec:method}

\vspace{-5pt}

In this section, we first introduce the overall dynamic architecture, and then detailed designs of the proposed multi-exit network and the multi-order difference detector are demonstrated. The supporting network components and training techniques will also be introduced.

\vspace{-10pt}

\subsection{Overview}


Figure \ref{fig_over} illustrates the overall architecture. Given a video sequence with $T$ frames, we aim to adaptively allocate subnets for different types of event boundaries. \ours{} is mainly composed of two key designs: the \textbf{multi-exit network} to enable adaptive inference at the video snippet level and the \textbf{multi-order difference detector} to achieve the dynamism by capturing the most distinctive motion patterns for boundary detection.

\vspace{-10pt}

\subsection{Multi-exit Network}

To achieve adaptive inference, we design a dynamic network with $L$ different boundary detectors, where these intermediate detectors are attached at varying depths of the image backbone (shown in the left plane of Figure \ref{fig_over}).

\vspace{-10pt}

\subsubsection{Shared Backbone and Separated Detectors.} Given the frame $\mathbf{I}_t$ at the timestamp $t$ and the centered local video snippet $\mathbf{V} = [\mathbf{I}_{t-k}, \ldots, \mathbf{I}_{t+k}]$, the output of the $l$-th detector ($l \in [1,\ldots,L]$) can be represented by:
\begin{equation}
    \mathbf{p}^l_t = h_l(f_l([\mathbf{I}_{t-k}, \ldots, \mathbf{I}_{t+k}], \theta_l), \phi_l),
\end{equation}
where $f_l$ is the subnet of backbone and $\theta_l$ denotes the partially shared parameters, $h_l$ is the specific boundary detector with parameter $\phi_l$ which is not shared across detectors. The combination of a shared backbone and separated detectors allows more targeted processing to various boundaries while still maintaining the efficiency property of multi-exit networks. $\mathbf{p}^l_t \in (0,1)$ denotes the boundary prediction score after the sigmoid function.

\vspace{-10pt}

\subsubsection{Multi-scale Spatio-temporal Features.} Following the common paradigm \cite{li2022structured,kang2022uboco}, $T$ frames are first sampled from the raw video, and sent into the image backbone (ResNet50 \cite{he2016deep} by default). Spatial features from different layers $[\mathbf{V}^1, \ldots, \mathbf{V}^L]$ are iteratively extracted during adaptive inference, severing as representations of different levels of semantics. For each scale, spatial pooling is then performed for each frame to save computations and obtain the 1d temporal feature. All temporal features from previous scales (if exist) are collected and concatenated to the current scale $l$ to obtain enriched feature $\mathbf{R}^l \in \mathbb{R}^{T \times l \times C}$.

\vspace{-15pt}



\subsubsection{Continuous Prediction.} Due to the ambiguous annotations for generic boundaries, we formulate GEBD as a global continuous prediction task supervised by soft labels \cite{li2022structured}. A much longer video sequence is sent to the network directly so that the huge repetitive computations of the backbone among adjacent overlapped windows can be saved. Due to the local property \cite{tan2023temporal,li2022structured} of event boundaries, we unfold the sequence into local windows for precise detection but on features after being extracted from the backbone. Specifically, $k$ frames before and after each candidate position in $\mathbf{R}^l$ are collected to form the local centered window of $t_w$ frames, where $t_w = 2k + 1$. The overall partitioned feature is $\mathbf{X}^l \in \mathbb{R}^{T \times l \times C \times t_w}$, which corresponds to the input of $l$-th attached boundary detector. Each local window corresponds to each timestamp and can produce a confidence score to represent the probability of being the boundary. Since the predictions are supervised by soft labels, the change of confidence scores between adjacent frames is smooth, we can then perform peak estimation to get boundaries $\mathbf{b}$:
\begin{equation}
    \mathbf{b} = \{t \mid \mathbf{p}_t > \mathbf{p}_{t-1}, \mathbf{p}_t > \mathbf{p}_{t+1}, \mathbf{p}_t > \epsilon \},
\end{equation}
where each timestamp corresponds to the local maximum that satisfies the classification threshold $\epsilon$ (e.g. 0.5) is considered as the boundary.

\vspace{-10pt}



\subsubsection{Partial Exit Criterion.} \label{sec:exit} We can easily determine whether a boundary candidate should exit at the $l$-th detector based on whether its prediction score exceeds a pre-defined threshold $\epsilon^l$. However, since the long video sequence we send into the network usually contains multiple boundaries with various characteristics, we can not simply follow existing multi-exit networks \cite{huang2018multi,yang2020resolution} to have the entire sequence exit at once. Therefore, we design a partial exit criterion from the global perspective on the whole sequence. Specifically, due to the locality of event boundaries, we only allow features around the detected point to exit and generate a mask $\mathbf{M}^l \in \{0,1\}^T$ indicating whether to drop or keep for features at each timestamp:
\begin{equation}
    \mathbf{M}^l_t = \mathbb{I}(t \notin [b-t_\mu^l, b+t_\mu^l]), \quad b \in \mathbf{b}, \quad t \in [1,\ldots,T],
\end{equation}
where $t_\mu^l$ determines the amount of exited features around the detected boundaries $\mathbf{b}$. Then we use the mask $\mathbf{M}^l$ to index which parts of the backbone features are kept:
\begin{equation}
    \mathbf{V}^{l+1} = \{f_l(\mathbf{V}_t) \mid \mathbf{M}^l_t=1\}.
\end{equation}

Since $t_\mu$ is adjustable, it controls the allocation of subnets to different samples as well as the trade-off between performance and computational costs. After partial exit, the remaining frame features are reorganized into a new sequence and fed into deeper backbone layers to obtain high-level semantics. Since the simply-merged features exhibit significant differences at the points where frames were previously exited and result in false boundaries, we pad the exited positions with repetitions of the nearby frame features to preserve the local information, and the sequence is padded back to the initial length $T$ before being sent to corresponding detectors. Furthermore, the boundaries detected within the padded frame features are also undesirable and will not be recorded.


Considering that the computational cost of the backbone network $f$ is significantly larger than that of the detector $h$ (over 10$\times$ in FLOPs), the efficiency gained from allowing more frames to exit early far outweighs the additional burden caused by hard-to-detect samples traversing all detectors.



\subsection{Multi-order Difference Detector}

The generic event boundary types span simple shot changes to complex sub-action changes. Besides the overall dynamic architecture, we further propose the multi-order difference detector utilizing coarse-to-fine features from the backbone and ensuring these various boundaries can be detected with specialized strategies. The detectors should explore distinctive representations among various boundaries and distinguish them to ensure the effectiveness of dynamism.
The detector based on multi-order temporal differences captures and amplifies the most significant motion pattern for boundary identification by modeling localized changes. From the right plane of Figure \ref{fig_over}, the detector is composed of a multi-order difference encoder (MDE) and a pairwise contrast module (PCM) for boundary identification and localization.

\vspace{-10pt}

\subsubsection{Multi-order Difference Encoder.} A transformer-liked architecture is proposed to capture multi-order temporal dependencies, which is composed of $n$ blocks with pooling layers among them.
The localized instantaneous change (i.e., the gradient of the time series) is crucial for boundary detection tasks \cite{tang2022progressive}. From Figure \ref{fig1}, the shot change has a clear discrete change point of its feature activation curve. For the sub-action change, the boundary information can only be distinguished after the 2nd temporal difference. In order to model various localized relationships, we propose the difference mixer (Diff Mixer in Figure \ref{fig_over}) with different choices of orders to capture the most distinctive motion patterns for various boundaries. Following ConvMixer \cite{trockman2022patches}, we first use small kernel convolutions to enable local interactions. Given a input tensor $\mathbf{X} \in \mathbb{R}^{T \times l \times C \times t_w}$, the mixer $g$ is:
\begin{equation}
    g(\mathbf{X}) = \mathtt{BN}(\sigma(\mathtt{DWConv}(\mathbf{X}))),
\end{equation}
where $\mathtt{BN}$ is batch normalization \cite{ioffe2015batch}, $\sigma$ is the non-linear activation fuction and $\mathtt{DWConv}$ is the depth-wise temporal convolution \cite{howard2017mobilenets}. To model the localized change pattern, we compute the first and second-order temporal differences to approximate multi-order derivatives. The final output of the proposed multi-order difference mixer can be written as:
\begin{equation}
    \label{eq_dx}
    \begin{aligned}
        \mathbf{X}_d &= \mathbf{X}_d^0 + \mathbf{X}_d^1 + \mathbf{X}_d^2 \\
        &= g(\mathbf{X}) + g(\Delta \mathbf{X}) + g(\Delta g(\Delta \mathbf{X})),
    \end{aligned}
\end{equation}
where $\Delta [\cdot]$ denotes the temporal difference for the input sequence. We decompose the complex instantaneous motion representation into its multi-order derivatives, revealing the crucial change pattern for better boundary identification. Based on different order choices and combinations in Eq. \ref{eq_dx}, the boundary information of candidate snippets with various temporal dependencies can be easily captured. We also replace the vanilla FFN with ConvFFN to improve robustness as observed in \cite{mao2022towards}. We chose the max pooling with stride 1 as the connection among blocks. Adjacent features are merged gradually to capture deep action-level representations without losing discriminativeness. The features from different blocks are concatenated as $\mathbf{D} \in \mathbb{R}^{T \times nl \times C \times t_w}$, which is the input of the following pairwise contrast module.

\vspace{-10pt}

\subsubsection{Pairwise Contrast Module.} Although the complex temporal dependencies have been decomposed into different levels through multi-order differences, a boundary can only be identified within the given video snippet, and it is still ambiguous to localize the specific timestamp where the change occurs. The representation of a video snippet centered by the candidate frame remains under-explored. We found that frames preceding and following the central frame should exhibit dissimilarity between groups but similarity within each group. In order to enhance the localization ability, we proposed a pairwise contrast module by utilizing the relationship of frames inside the given snippet. Specifically, we calculate the frame-level pairwise similarity as below:
\begin{equation}
    \mathbf{S}_t(i,j) = \mathtt{Sim}(\mathbf{D}_t^i, \mathbf{D}_t^j), \quad i,j \in [1,\ldots,t_w], \quad t \in [1,\ldots,T],
\end{equation}
where $\mathbf{S}_t \in \mathbb{R}^{nl \times t_w \times t_w}$ is the similarity map of $n$ features from different blocks in MDE, $\mathbf{D}_i^t$ and $\mathbf{D}_j^t$ are features inside the given snippet and $\mathtt{Sim}$ is the cosine similarity function. Since the crucial boundary information relies on the local diagonal patterns in $\mathbf{S}$ \cite{kang2022uboco}, the obtained similarity map is then sent into a light-weighted encoder composed of a few layers of convolutions for feature refinement.

Lastly, we fuse the outputs of two branches in the proposed detector through cross SE (squeeze-and-excitation) \cite{hu2018squeeze} modules to re-weighting and balancing the features. The fused features are passed through MLP-based classifiers to generate the final predictions.

\vspace{-10pt}

\subsection{Training Details}

\subsubsection{Gaussian Smoothing.} As mentioned before, we follow the continuous paradigm that each frame-wise prediction is merged to obtain the scores $\mathbf{p}$ with length $T$. Since the annotations for GEBD are subjective and ambiguous, directly using these hard labels to optimize the network could lead to poor generalization ability. Therefore, we smoothed the one-hot labels with a Gaussian kernel to generate soft labels $\widetilde{\mathbf{y}}$. The window size of the Gaussian kernel remains the same as the length of the video snippet and $\sigma=1$ in all experiments.

\vspace{-15pt}

\subsubsection{Loss Function.} After obtaining the predictions from detectors, we calculate the binary cross entropy loss for each one. The total training loss can be written as:
\begin{equation}
    \mathcal{L} = \sum_l{\alpha_l[-\widetilde{\mathbf{y}}_l\log(\mathbf{p}_l) - (1 - \widetilde{\mathbf{y}}_l)\log(1 - \mathbf{p}_l)]},
\end{equation}
where losses of $L$ detectors are summed together by weight $\alpha$ to balance the optimization of different subnets.


\section{Experiment}
\label{sec:exp}


\subsection{Experimental Settings}


\subsubsection{Datasets.} We perform evaluations on Kinetics-GEBD \cite{shou2021generic} and TAPOS \cite{shao2020intra} datasets following others. Kinetics-GEBD is a large-scale dataset comprising 54,691 videos and 1,290,000 temporal boundaries, designed to annotate generic event boundaries in the wild. The ratio of training, validation, and testing videos of Kinetics-GEBD is nearly 1:1:1. Each video is annotated by 5 annotators with an average of 4.77 boundaries. Due to the annotation of the test set being publicly unavailable, we trained our network on the training set and evaluated it on the validation set. The TAPOS dataset contains 21 different actions in Olympic sports videos. There are 13,094 action instances for training and 1,790 instances for validation. The boundaries in TAPOS are more fine-grained and challenging to detect. Following \cite{shou2021generic}, we re-purpose TAPOS by trimming each action instance with its action label hidden and conducting experiments on them.

\vspace{-15pt}

\subsubsection{Evaluation Protocol.} We employ the F1 score as the measurement for GEBD. As mentioned in \cite{shou2021generic}, the Relative Distance (Rel.Dis.) represents the error between the detected and ground truth timestamps divided by the length of the corresponding action instance. Each detection result is compared with the annotations from each rater, and the highest F1 score is considered the final outcome. We present F1 scores across various Rel.Dis. thresholds ranging from 0.05 to 0.5, with an increment of 0.05. We mainly compare F1@0.05 with others, which keep the same evaluation protocol as the official \href{https://sites.google.com/view/loveucvpr23/track1}{CVPR'23 LOVEU challenge}.

\vspace{-15pt}

\subsubsection{Implementation Details.} In practise, we choose ResNet50 \cite{he2016deep} pretained on ImageNet \cite{deng2009imagenet} as the backbone for a fair comparison. Since the video duration and fps are similar in Kinetics-GEBD, we uniformly sample 100 frames (i.e., $T=100$) from each video as the inputs. The TAPOS has a large variety of duration over instances from a few seconds to around 5 minutes. Following \cite{tan2023temporal}, we split the instances without overlapping and sample 100 frames by keeping a similar fps to Kinetics-GEBD. The scores of sub-instances are merged together to generate the final prediction. The $k$ is set to 8, and the length of the local window $t_w$ is 17. We choose the last 3 layers of ResNet50 for adaptive inference, and the number of blocks $n$ in the transformer-liked encoder is 3. The cosine function is chosen as the measurement of the pairwise similarity. The whole model is trained end-to-end for 20 epochs with Adam \cite{kingma2014adam} and a base learning rate of 1e-2.

\vspace{-10pt}

\subsection{Main Results}

\begin{table*}[t]
\setlength{\tabcolsep}{5pt}
\centering
\caption{Comparisons in terms o F1 score (\%) on Kinetics-GEBD with Rel.Dis. threshold from 0.05 to 0.5. \textsuperscript{\dag}: CSN \cite{tran2019video} backbone. 
}
\label{tab:exp_kgebd}
\rowcolors{14}{gray!10}{gray!10}
\resizebox{\textwidth}{!}{
\begin{tabular}{l|c|cccccccccc}
\toprule
\multirow{2}{*}{Method} & \multicolumn{11}{c}{F1 @ Rel. Dis.}                                                                                                                                           \\ \cline{2-12} 
                        & 0.05          & 0.1           & 0.15          & 0.2           & 0.25          & 0.3           & 0.35          & 0.4           & 0.45          & 0.5           & avg           \\ \midrule
BMN \cite{lin2019bmn}                    & 18.6          & 20.4          & 21.3          & 22.0          & 22.6          & 23.0          & 23.3          & 23.7          & 23.9          & 24.1          & 22.3          \\
BMN-StartEnd \cite{lin2019bmn}           & 49.1          & 58.9          & 62.7          & 64.8          & 66.0          & 66.8          & 67.4          & 67.8          & 68.1          & 68.3          & 64.0          \\
TCN-TAPOS \cite{lea2016segmental}              & 46.4          & 56.0          & 60.2          & 62.8          & 64.5          & 65.9          & 66.9          & 67.6          & 68.2          & 68.7          & 62.7          \\
TCN \cite{lea2016segmental}                    & 58.8          & 65.7          & 67.9          & 69.1          & 69.8          & 70.3          & 70.6          & 70.8          & 71.0          & 71.2          & 68.5          \\
PC \cite{shou2021generic}                     & 62.5          & 75.8          & 80.4          & 82.9          & 84.4          & 85.3          & 85.9          & 86.4          & 86.7          & 87.0          & 81.7          \\
SBoCo-Res50 \cite{kang2022uboco}            & 73.2          & -             & -             & -             & -             & -             & -             & -             & -             & -             & 86.6          \\
Temporal Perceiver \cite{tan2023temporal}     & 74.8          & 82.8          & 85.2          & 86.6          & 87.4          & 87.9          & 88.3          & 88.7          & 89.0          & 89.2          & 86.0          \\
CVRL \cite{li2022end}                   & 74.3          & 83.0          & 85.7          & 87.2          & 88.0          & 88.6          & 89.0          & 89.3          & 89.6          & 89.8          & 86.5          \\
DDM-Net \cite{tang2022progressive}                & 76.4          & 84.3          & 86.6          & 88.0          & 88.7          & 89.2          & 89.5          & 89.8          & 90.0          & 90.2          & 87.3          \\
SC-Transformer \cite{li2022structured}         & 77.7          & 84.9          & 87.3          & 88.6          & 89.5          & 90.0          & 90.4          & 90.7          & 90.9          & 91.1          & 88.1          \\ \midrule
\textbf{DyBDet (ours)}  & \textbf{79.6} & \textbf{85.8} & \textbf{88.0} & \textbf{89.3} & \textbf{90.1} & \textbf{90.7} & \textbf{91.1} & \textbf{91.5} & \textbf{91.7} & \textbf{91.9} & \textbf{89.0} \\
\textbf{DyBDet (ours)}\textsuperscript{\dag} & \textbf{83.1}  & \textbf{88.4}  & \textbf{90.4}  & \textbf{91.3}  & \textbf{92.0}  & \textbf{92.5}  & \textbf{92.7}  & \textbf{93.0}  & \textbf{93.2} & \textbf{93.3}  & \textbf{91.0} \\  \bottomrule
\end{tabular}
}
\end{table*}


\begin{table*}[t]
\setlength{\tabcolsep}{5pt}
\centering
\caption{Comparison with others in terms of F1 score (\%) on TAPOS with Rel.Dis. threshold from 0.05 to 0.5 with 0.05 interval.}
\label{tab:exp_tapos}
\rowcolors{12}{gray!10}{gray!10}
\resizebox{\textwidth}{!}{
\begin{tabular}{l|c|cccccccccc}
\toprule
\multirow{2}{*}{Method} & \multicolumn{11}{c}{F1 @ Rel. Dis.}                                                                                                                                                                                                                                                                    \\ \cline{2-12} 
                        & 0.05                     & 0.1                      & 0.15                     & 0.2                      & 0.25                     & 0.3                      & 0.35                     & 0.4                      & 0.45                     & 0.5                      & avg                      \\ \midrule
ISBA \cite{ding2018weakly}                   & 10.6                     & 17.0                     & 22.7                     & 26.5                     & 29.8                     & 32.6                     & 34.8                     & 36.9                     & 38.2                     & 39.6                     & 30.2                     \\
TCN \cite{lea2016segmental}                    & 23.7                     & 31.2                     & 33.1                     & 33.9                     & 34.2                     & 34.4                     & 34.7                     & 34.8                     & 34.8                     & 34.8                     & 64.0                     \\
CTM \cite{huang2016connectionist}                    & 24.4                     & 31.2                     & 33.6                     & 35.1                     & 36.1                     & 36.9                     & 37.4                     & 38.1                     & 38.3                     & 38.5                     & 35.0                     \\
TransParser \cite{shao2020intra}            & 28.9                     & 38.1                     & 43.5                     & 47.5                     & 50.0                     & 51.4                     & 52.7                     & 53.4                     & 54.0                     & 54.5                     & 47.4                     \\
PC \cite{shou2021generic}                     & 52.2                     & 59.5                     & 62.8                     & 64.6                     & 65.9                     & 66.5                     & 67.1                     & 67.6                     & 67.9                     & 68.3                     & 64.2                     \\
Temporal Perceiver \cite{tan2023temporal}     & 55.2 & 66.3 & 71.3 & 73.8 & 75.7 & 76.5 & 77.4 & 77.9 & 78.4 & 78.8 & 73.2 \\
DDM-Net \cite{tang2022progressive}                & 60.4                     & 68.1                     & 71.5                     & 73.5                     & 74.7                     & 75.3                     & 75.7                     & 76.0                     & 76.3                     & 76.7                     & 72.8                     \\
SC-Transformer \cite{li2022structured}         & 61.8 & 69.4 & 72.8 & 74.9 & 76.1 & 76.7 & 77.1 & 77.4 & 77.7 & 78.0 & 74.2 \\ \midrule
\textbf{DyBDet (ours)}  & \textbf{62.5}            & \textbf{70.1}            & \textbf{73.4}            & \textbf{75.6}            & \textbf{76.7}            & \textbf{77.2}            & \textbf{77.5}            & \textbf{77.9}            & \textbf{78.1}            & \textbf{78.4}            & \textbf{74.7}            \\ \bottomrule
\end{tabular}
}
\vspace{-15pt}
\end{table*}

\begin{figure}
    \centering
    \includegraphics[width=0.6\linewidth]{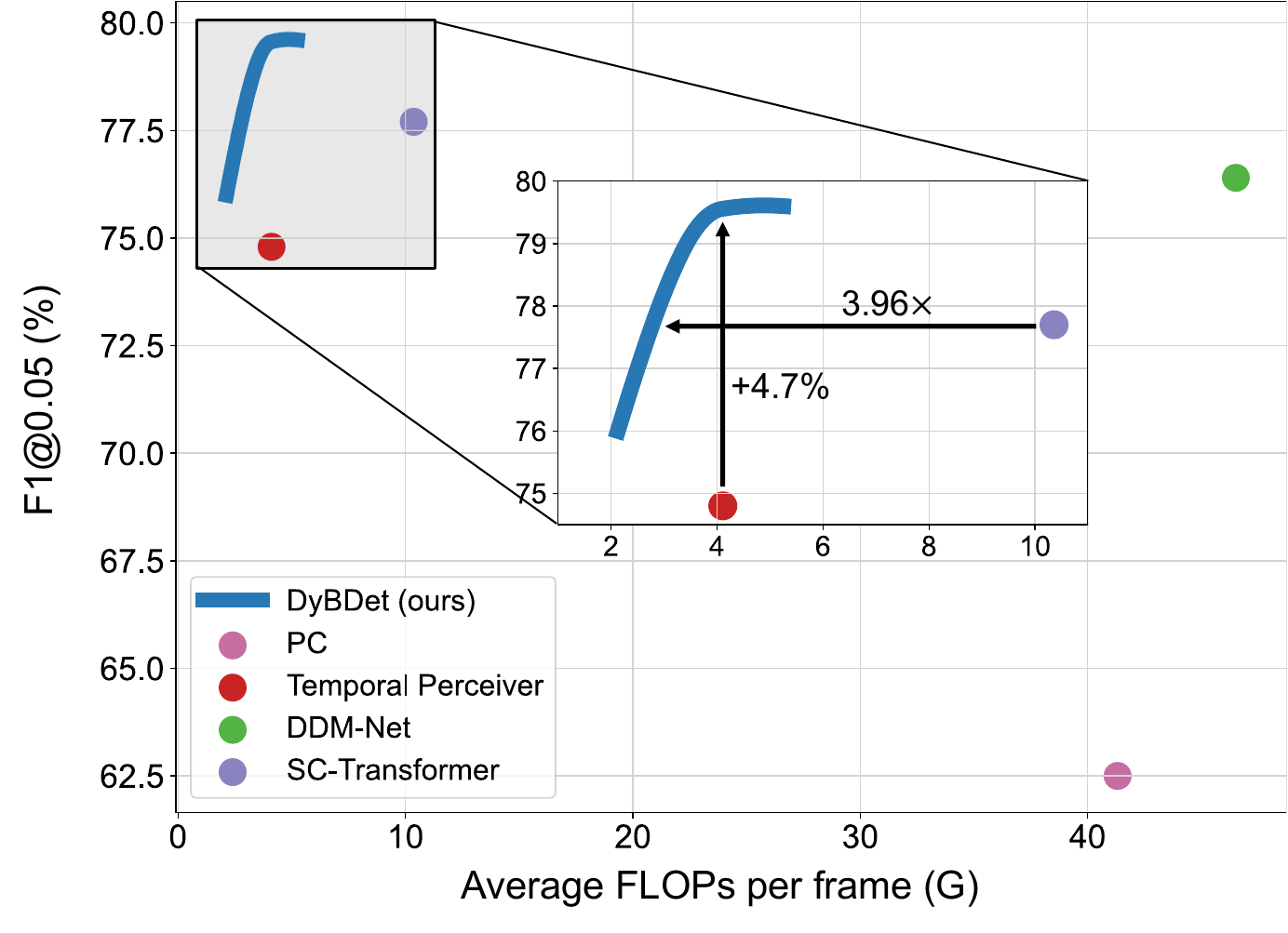}
    \caption{Comparisons of F1@0.05 v.s. FLOPs on Kinetics-GEBD with others. We report the average FLOPs per frame through the whole inference pipeline.}
    \label{fig_flops}
\end{figure}



\subsubsection{Kinetics-GEBD.} We evaluated the proposed \ours{} on the validation set of Kinetics-GEBD \cite{shou2021generic} in Table \ref{tab:exp_kgebd}. Our approach significantly outperforms all counterparts, particularly under the most strict Rel.Dis. constraint. Specifically, our method achieves improvements of 1.9\% and 0.9\% in F1@0.05 and average compared to the current sota, respectively. Moreover, when equipping with a much more powerful CSN \cite{tran2019video} backbone, we can achieve extra improvement significantly, demonstrating \ours{} is not limited by the backbone choices. We also provide the results of computational costs (measured in FLOPs) in Figure \ref{fig_flops}. Since our method is \textbf{dynamic}, we can adjust the amount of exited features $t_\mu$ to control the computational budgets and obtain a continuous curve standing for adaptive inference, indicating the flexibility of the proposed method in practical deployment with different computational constraints. As the frames are only inferred once within the backbone, we performed a 3.2\% improvement with 11.3$\times$ fewer FLOPs compared to DDM-Net \cite{tang2022progressive}. Notably, our proposed \ours{} achieved a significant improvement of 4.7\% compared to Temporal Perceriver \cite{tan2023temporal} under the same computational budget, and 3.96$\times$ fewer FLOPs compared to SC-Transformer \cite{li2022structured} under the same detection performance. We also found that if all frame features are exited at the first detector, the detection performance is already comparable while requiring much fewer computations (around 2 GFLOPs per frame), highlighting the effectiveness of the proposed method.

\vspace{-15pt}

\subsubsection{TAPOS.} We also conduct experiments on TAPOS \cite{shao2020intra} in Table \ref{tab:exp_tapos}. \ours{} also achieves state-of-the-art and increases the performance by 0.7\% and 0.5\% in F1@0.05 and average, respectively. Since TAPOS contains more fine-grained action instances than Kinetics-GEBD and with a much larger variety of video durations, the results also demonstrate the scalability of the proposed \ours{}.

\vspace{-10pt}

\subsection{Ablation Study}

In this section, we aim to study the individual contributions of each component to form the proposed dynamic architecture. All ablations are conducted on the Kinetics-GEBD dataset with a fixed random seed for training.

\begin{table}[t]
\setlength{\tabcolsep}{2pt}
  \begin{minipage}[b]{0.29\linewidth}
    \centering
    \caption{Study on modules in the detector.}
    \label{tab_det}
    \rowcolors{6}{gray!10}{gray!10}
    \resizebox{1.0\linewidth}{!}{
    \begin{tabular}{ccc|cccc}
        \toprule
        PCM        & MDE         & Fuse       & 0.05     & avg.     \\ \midrule
                   &             &            & 52.6     &  65.5    \\
                   & \checkmark  &            & 73.2     &  85.0    \\
        \checkmark &             &            & 76.3     &  87.2    \\
        \checkmark & \checkmark  &            & 78.5     &  88.4    \\ \midrule
        \checkmark & \checkmark  & \checkmark & \textbf{79.6}  &  \textbf{89.0}     \\ \bottomrule
        \end{tabular}
    }
  \end{minipage}\hfill
  \begin{minipage}[b]{0.34\linewidth}
    \centering
    \caption{Study on design variations in MDE.}
    \label{tab_diff}
    \rowcolors{6}{gray!10}{gray!10}
    \resizebox{1.0\linewidth}{!}{
    \begin{tabular}{cc|c|cc}
        \toprule
        Mixer & \#blk & Param. & 0.05 & avg.  \\ \midrule
        SA & 3 & 27.3 & 77.6 & 88.0 \\
        Conv1d & 3 & 29.7 & 78.2 & 88.3 \\
        Conv1d & 6 & 37.0 & 78.5 & 88.4 \\
        M-Diff. & 6 & 27.6 & 78.2 & 88.4 \\ \midrule
        M-Diff. & 3 & 22.5 & \textbf{79.6} & \textbf{89.0} \\ \bottomrule
        \end{tabular}
    }
  \end{minipage}\hfill
  \begin{minipage}[b]{0.33\linewidth}
    \centering
    \caption{Composition of difference order choices.}
    \label{tab_order}
    \rowcolors{6}{gray!10}{gray!10}
    \resizebox{1.0\linewidth}{!}{
    \begin{tabular}{lll|cc}
        \toprule
        Det.1       & Det.2      & Det.3     & 0.05     & avg.  \\ \midrule
        $\mathit{0}$ & $\mathit{0}$ & $\mathit{0}$   &  76.3   & 87.2     \\
        +$\mathit{1}$ & +$\mathit{1}$ & +$\mathit{1}$   &  78.5   & 88.5    \\
        +$\mathit{1}$+$\mathit{2}$ & +$\mathit{1}$+$\mathit{2}$ & +$\mathit{1}$+$\mathit{2}$   & 78.9   & 88.7  \\ \midrule
        \multicolumn{3}{c|}{learned}  & 78.2  & 88.3   \\ \midrule
        $\mathit{0}$ & +$\mathit{1}$ & +$\mathit{1}$+$\mathit{2}$  & \textbf{79.6}  & \textbf{89.0}   \\ \bottomrule
        \end{tabular}
    }
  \end{minipage}
  \vspace{-10pt}
\end{table}

\vspace{-12pt}

\subsubsection{Multi-order Difference Detector.} To achieve adaptive processing in the dynamic architecture, we need a detector that can distinguish and capture the most representative features of various types of boundaries. The proposed multi-order difference detector is composed of the multi-order difference encoder (MDE) and the pairwise contrast module (PCM). The former adaptively models temporal differences for boundary identification, and the latter encompasses pairwise similarities to enhance the boundary representation. We first study how each module contributes to the detector, and the results are shown in Table \ref{tab_det}. Similar to the conclusions from \cite{kang2022uboco} and \cite{tang2022progressive}, the similarity map captured from the pairwise contrast module (PCM) greatly improves the detection performance. The localized change pattern along the temporal dimension from PCM is regarded as a significant boundary representation, which is intuitive and effective. Considering the variations of temporal dependencies of different types of boundaries, we further enhance the boundary features with the multi-order difference encoder (MDE). The naive per-frame features are replaced by the multi-order temporal differences, enabling the capability of modeling complex changes as well as the distinguishment for different kinds of boundaries, leading to further performance gains of the dynamic architecture from 76.3\% to 78.5\%. Since the proposed MDE is composed of several transformer-liked layers and gradually merges boundary features, the output can also contribute to boundary detection. Simply fusing the outcomes from two modules can obtain the best result.

\vspace{-12pt}

\subsubsection{Temporal Differences.} To further verify the performance is not gained from in-depth temporal feature extraction in MDE, we replace the multi-order temporal differences (M-Diff.) in the token mixer with more parameterized operations and also increase the number of blocks. From Table \ref{tab_diff}, simply adopting networks with more parameters for temporal modeling does not yield higher performance, indicating the importance of multi-order temporal difference for boundary detection. Moreover, computing differences in deep temporal features can also be harmful since the features of adjacent frames are becoming similar \cite{shi2023tridet}.

\vspace{-12pt}


\begin{figure}[tb]
    \centering
    \begin{minipage}[h]{0.38\linewidth}
        \centering
        \includegraphics[width=1\linewidth]{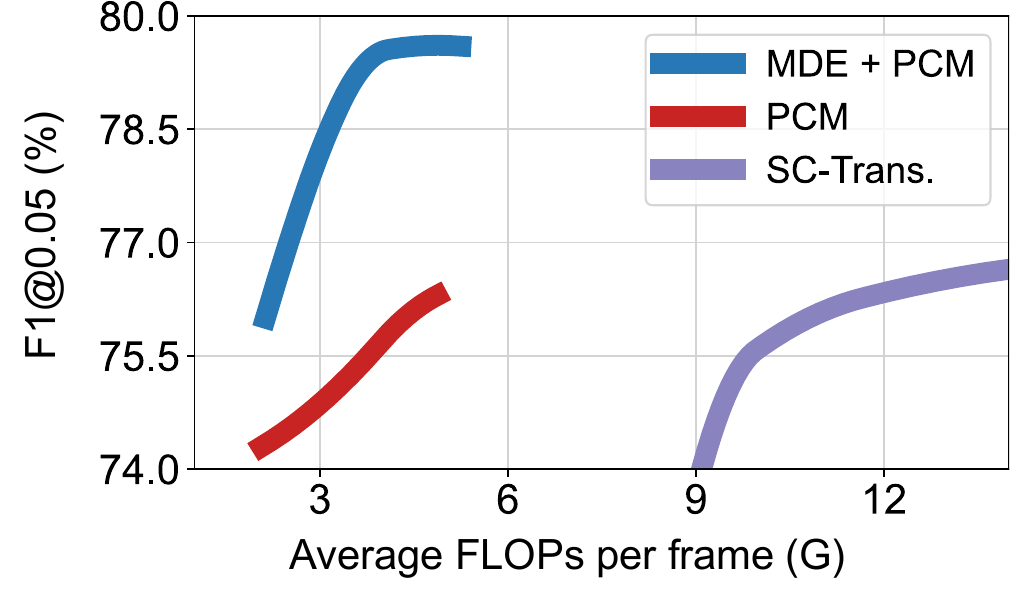}
        \vspace{-10pt}
        \figcaption{Dynamic networks with partial exit composed by different detectors. MDE+PCM indicates the detector in \ours{}.}
        \vspace{-10pt}
        \label{fig_dy}
    \end{minipage}
    \hfill
    \begin{minipage}[h]{0.59\linewidth}
        \centering
        \includegraphics[width=1\linewidth]{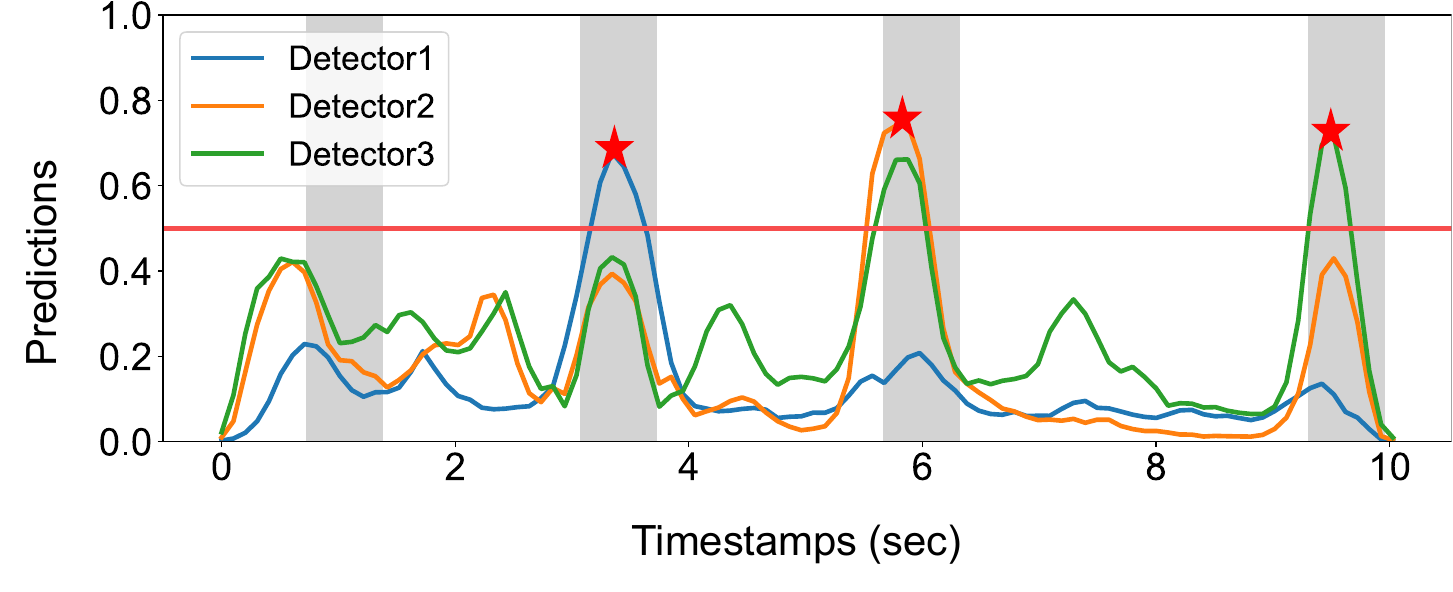}
        \vspace{-15pt}
        \figcaption{The predictions of different detectors \textbf{w/o partial exit}. The red line represents the threshold $\epsilon$, and the gray area indicates the ground-truth labels w.r.t F1@0.05. Stars are detected boundaries.}
        \vspace{-10pt}
        \label{fig_3heads}
    \end{minipage}
\end{figure}

\subsubsection{Affinity with Dynamic Architecture.} Besides the capability of boundary detection, the proposed multi-order difference detector is designed to ensure various boundaries can be effectively detected with different strategies. To study the affinity with the dynamic architecture, we replace the detectors attached at different depths of the backbone with other alternatives and perform adaptive inference in Figure \ref{fig_dy}. From the results, although the detector in SC-Transformer \cite{li2022structured} can achieve high performance alone, its application to the dynamic network suffers from poor generalizations due to the heaviness. The performance even drops 1.9\% under the same computational budget when becoming dynamic. We also examine the detector composed of PCM alone, which can achieve comparable performance as discussed in Table \ref{tab_det} while being extremely lightweight and efficient. However, we can not witness performance saturation when given more computational budgets, indicating that almost all boundaries favor the deepest subnet. As PCM fails to effectively distinguish various boundaries, the adaptive inference of the corresponding dynamic network loses its intended significance. Different from them, since the proposed detector can adaptively explore the most distinctive features of boundaries through multi-order differences, \ours{} can effectively detect various boundaries with different subnets and already achieve the best performance without traversing the whole network for all candidates. We also plot the predictions w.r.t subnets corresponding to different detectors in Figure \ref{fig_3heads}, and each subnet can partially find specific boundaries. If these boundaries are aggregated by the multi-exit network, we can obtain the best overall performance, indicating the effectiveness of adaptive inference.

\begin{table}[t]
\setlength{\tabcolsep}{2.5pt}
  \begin{minipage}[b]{0.35\linewidth}
    \centering
    \caption{Study on the partial exit criterion.}
    \label{tab_partial}
    \rowcolors{5}{gray!10}{gray!10}
    \resizebox{1.0\linewidth}{!}{
    \begin{tabular}{l|cc}
        \toprule
        Strategy & F1@0.05 & GFLOPs \\ \midrule
        Post Combine & 78.1  & 5.25 \\
        Exit w/o Pad & 78.4 & 4.03 \\
        Exit LinearPad & 79.0 & 4.27 \\ \midrule
        Exit RepeatPad & \textbf{79.6} & 4.15 \\ \bottomrule
        \end{tabular}
    }
  \end{minipage}\hfill
  \begin{minipage}[b]{0.33\linewidth}
    \centering
    \caption{Comparisons on inference latency (ms).}
    \label{tab_latency}
    \rowcolors{5}{gray!10}{gray!10}
    \resizebox{1.0\linewidth}{!}{
    \begin{tabular}{l|ccc}
        \toprule
        Method & F1@0.05 & latency \\ \midrule
        PC \cite{shou2021generic} & 62.5 & 5.32 \\
        DDM-Net \cite{tang2022progressive} & 76.4 & 25.75 \\
        SC-Trans. \cite{li2022structured} & 77.7 & 1.03 \\ \midrule
        DyBDet & \textbf{79.6} & \textbf{0.84} \\ \bottomrule
        \end{tabular}
    }
  \end{minipage}\hfill
  \begin{minipage}[b]{0.29\linewidth}
    \centering
    \caption{Study on different local window sizes.}
    \label{tab_win}
    \rowcolors{4}{white}{gray!10}
    \resizebox{1.0\linewidth}{!}{
    \begin{tabular}{c|cc}
        \toprule
        \#win $t_w$ & F1@0.05 & F1@avg \\ \midrule
        9 & 78.0 & 88.2 \\
        13 & 78.6 & 88.6 \\
        \textbf{17} & \textbf{79.6} & \textbf{89.0} \\
        21 & 79.5 & 89.0 \\
        \bottomrule
    \end{tabular}
    }
  \end{minipage}
  \vspace{-15pt}
\end{table}

\vspace{-15pt}

\subsubsection{Order Choices.} To further study how the proposed detector achieves adaptive inference, we provide the results of different choices of temporal difference orders for the three detectors (Det.1/2/3) attached at varying depths of the backbone in Table \ref{tab_order}. We found that although high-order differences can achieve performance gain, adding it for all detectors equally is worse than the choices of different order combinations. Intuitively, boundaries like shot changes can be easily detected through pairwise contrast in PCM, adding additional temporal differences could introduce noises and vanish the most distinctive boundary representations. We also provide the results that the computed orders are summed together through learnable parameters, yet the performance drops by 1.4\%. Since the image backbone is trained together with detectors, letting samples with complex temporal dependencies exit early could do harm to the training of the backbone as early layers lose sensitivities to capture low-level appearance information, which could be essential. The findings demonstrate the separated order choices in \ours{} can benefit the joint training of the supernet and boost the overall performance.

\vspace{-15pt}

\subsubsection{Partial Exit Criterion.} We propose the partial exit criterion to enable dynamism, and we study its contribution in Table \ref{tab_partial}. We first replace the partial exit strategy by simply combining the detection results from different detectors (Post Combine). However, this leads to a 1.5\% decrease in performance along with additional computational overhead. The partial exit criterion makes the detection of different boundaries distinguishable, indicating the effectiveness of dynamic detection for GEBD. We also study different padding methods as discussed in section \ref{sec:exit}. Padding the previously exited timestamps can mitigate their influence on subsequent detections. Moreover, using repetitions of nearby remaining frame features for padding outperforms linear interpolation.

\vspace{-15pt}

\subsubsection{Inference Latency.} In addition to the efficiency results w.r.t FLOPs, we also provide the practical inference latency on the GeForce RTX 4090 GPU. From Table \ref{tab_latency}, the proposed dynamic method can achieve high performance with practical efficiency without customized CUDA kernels.

\vspace{-15pt}

\subsubsection{Window Size.} We also conduct ablations on different lengths of the local window size for detectors in Table \ref{tab_win}. Given an input sequence of length $T$, the larger window size $t_w$ means a larger temporal respective field for boundary detection. However, due to the locality of boundaries, increasing the window size does not necessarily bring performance growth while with huge extra computations.

\begin{figure}[t]
    \centering
    \includegraphics[width=0.75\linewidth]{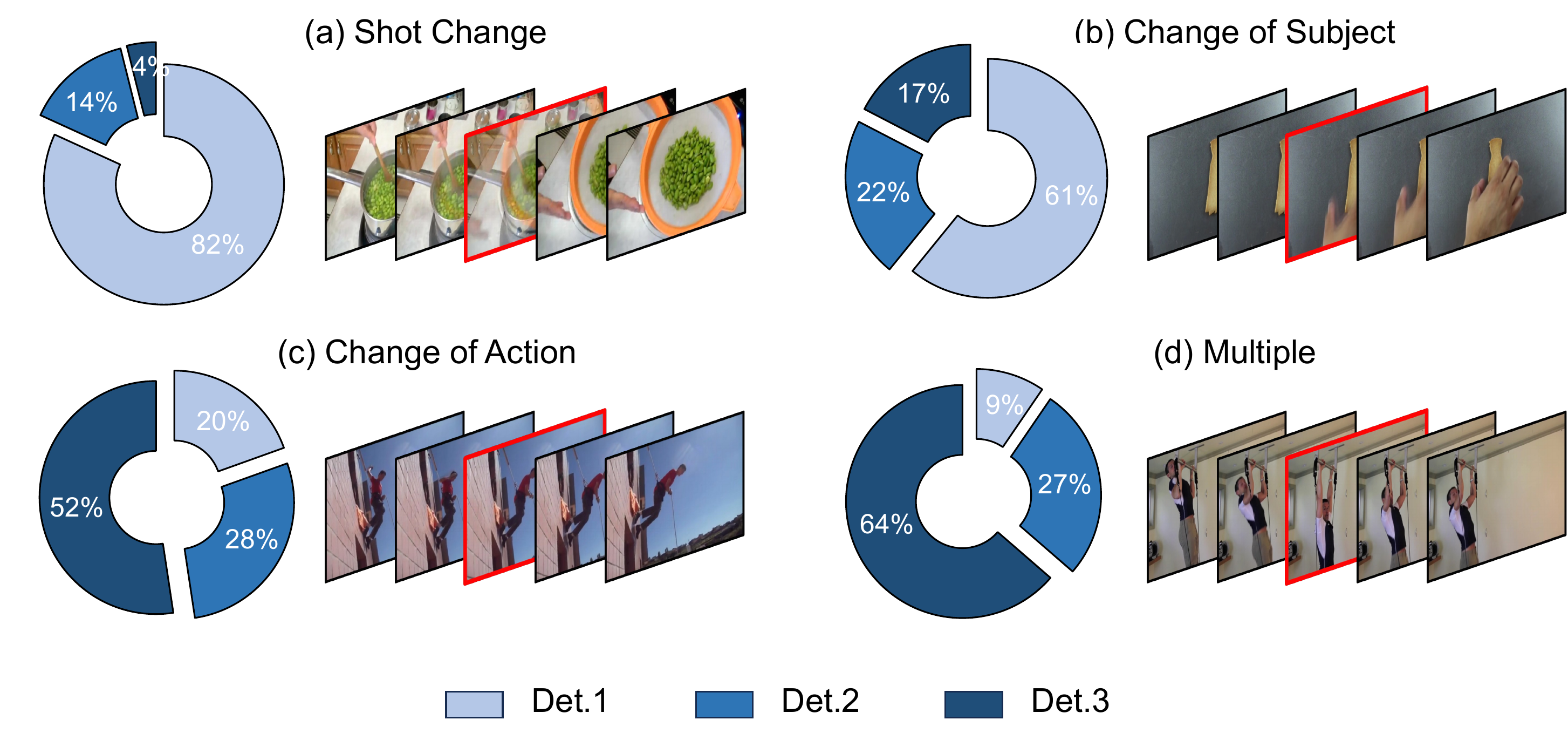}
    \caption{Class-specific distribution of True Positive samples under Rel.Dis. 0.05 on different detectors.}
    \label{fig_pie}
    \vspace{-15pt}
\end{figure}

\begin{figure}[t]
    \centering
    \includegraphics[width=0.8\linewidth]{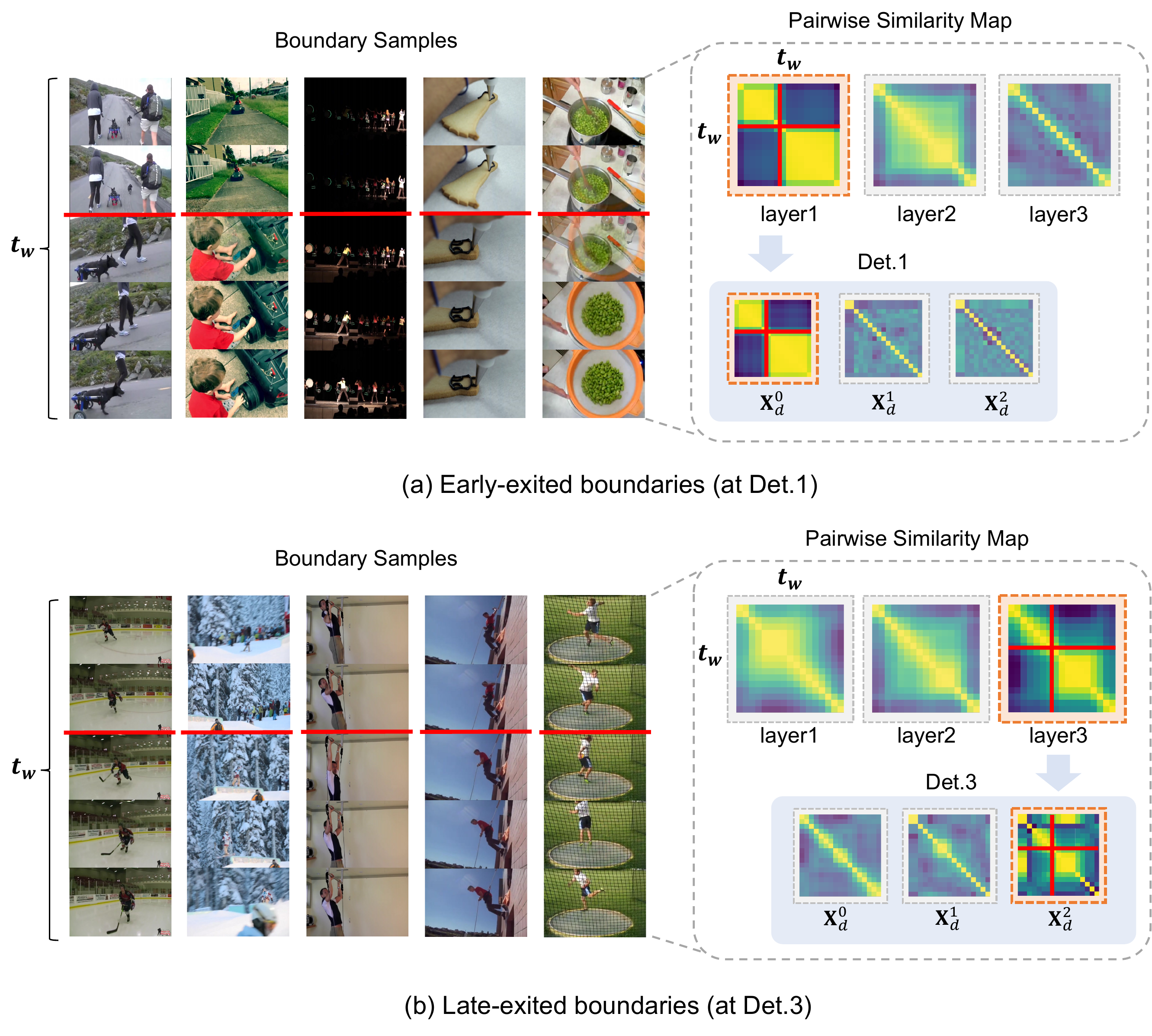}
    \caption{Visualizations of the pairwise similarity maps vary detectors and computed orders. Red lines indicate the boundaries. (a): the early-exited samples have clear boundary information only based on the lowest level features from \textit{layer1} and the zero-order difference $\textbf{X}_d^0$. (b): the late-exited samples have more discriminate characteristics based on high-level features from \textit{layer3} and the second-order difference $\textbf{X}_d^2$.}
    \label{fig_diffmap}
    \vspace{-15pt}
\end{figure}

\vspace{-10pt}

\subsection{Visualization and Discussion}

\subsubsection{Class-specific Statistics.} We analyze the class-specific statistics in Figure \ref{fig_pie} to understand how \ours{} achieves fine-grained detection for generic event boundaries. Note that although class-specific annotations (like shot changes) are available in Kinetics-GEBD, we treat them as binary labels during training and evaluation as others and only use them for in-depth study. We choose the top four categories with the highest quantity and report the distribution of detection results on different detectors for each class. Since only binary labels are available in predictions, we are unable to compute a confusion matrix for each category, so we count the number of True Positive (TP) samples as an alternative metric. For cases with large appearance-level shifts like shot changes and dominate subject changes, they favor shallow subnets and tend to exit early. On the other hand, identifying action changes and multiple movements requires deeper and more complex spatiotemporal modeling. As a result, generic boundaries with various characteristics can be automatically detected with the most suitable subnet, achieving high overall performance and efficiency.

\vspace{-10pt}

\subsubsection{Pairwise Similarity Map.} We present examples with $t_w$ frames in Figure \ref{fig_diffmap} and visualize their pairwise similarity maps of $t_w \times t_w$ captured by detectors with different orders of temporal difference. The red lines present the ground-truth boundaries. The pairwise similarity maps are then sent to the contrast module to amplify the discriminatives. Therefore, the diagonal pattern (similarity within the same side of frame groups and dissimilarity between groups) is essential to maximizing the boundary information as discussed in \cite{kang2022uboco}. From the left panel of Figure \ref{fig_diffmap}, samples like shot changes have distinctive background changes rather than the main objects. These low appearance-level features are only preserved in the shallow layers of the backbone. As a result, we can only distinguish a clear boundary based on the features from \textit{layer1}. Moreover, these characteristics also have fewer temporal dependencies, mainly captured from the zero-order difference of Det.1. On the other hand, the samples like sub-action changes on the right panel require more fine-grained features from deeper layers of the backbone (\textit{layer3}), as well as complex temporal change pattern modeling to identify the ambiguous boundary. The crucial diagonal patterns are found mainly in the second-order difference of Det.3, indicating high-level features with high-order temporal differences are needed. Overall, the visualizations demonstrate the demands of adaptive processing and the effectiveness of \ours{}.

\vspace{-5pt}

\section{Conclusion}

\vspace{-5pt}

In this paper, we present a dynamic model for generic event boundary detection (GEBD) named \ours{}. Unlike existing methods that detect various boundaries with the same protocol, we aim to dynamically allocate subnets to different video snippets based on their distinctive features, outperforming the previous methods by a large margin. As for \textbf{future works}, we plan to evaluate the generalizability of our method on more tasks with temporal localization.

\vspace{-10pt}

\subsubsection{Acknowledgement.} This work was supported in part by the National Science and Technology Major Project under Grant 2022ZD0115803, in part by the Natural Science Basic Research Plan in Shaanxi Province of China under Grant 2023-JC-JQ-51, and in part by the National Natural Science Foundation of China under Grants 62206215.

%
%
\bibliographystyle{splncs04}
\bibliography{egbib}

\begin{thebibliography}{10}
\providecommand{\url}[1]{\texttt{#1}}
\providecommand{\urlprefix}{URL }
\providecommand{\doi}[1]{https://doi.org/#1}

\bibitem{aakur2019perceptual}
Aakur, S.N., Sarkar, S.: A perceptual prediction framework for self supervised event segmentation. In: CVPR. pp. 1197--1206 (2019)

\bibitem{alayrac2017joint}
Alayrac, J.B., Laptev, I., Sivic, J., Lacoste-Julien, S.: Joint discovery of object states and manipulation actions. In: ICCV. pp. 2127--2136 (2017)

\bibitem{chen2019you}
Chen, Z., Li, Y., Bengio, S., Si, S.: You look twice: Gaternet for dynamic filter selection in cnns. In: CVPR. pp. 9172--9180 (2019)

\bibitem{cheng2022tallformer}
Cheng, F., Bertasius, G.: Tallformer: Temporal action localization with a long-memory transformer. In: ECCV. pp. 503--521. Springer (2022)

\bibitem{dai2021dynamic}
Dai, X., Chen, Y., Xiao, B., Chen, D., Liu, M., Yuan, L., Zhang, L.: Dynamic head: Unifying object detection heads with attentions. In: CVPR. pp. 7373--7382 (2021)

\bibitem{deng2009imagenet}
Deng, J., Dong, W., Socher, R., Li, L.J., Li, K., Fei-Fei, L.: Imagenet: A large-scale hierarchical image database. In: CVPR. pp. 248--255. Ieee (2009)

\bibitem{ding2018weakly}
Ding, L., Xu, C.: Weakly-supervised action segmentation with iterative soft boundary assignment. In: CVPR. pp. 6508--6516 (2018)

\bibitem{farha2019ms}
Farha, Y.A., Gall, J.: Ms-tcn: Multi-stage temporal convolutional network for action segmentation. In: CVPR. pp. 3575--3584 (2019)

\bibitem{gygli2018ridiculously}
Gygli, M.: Ridiculously fast shot boundary detection with fully convolutional neural networks. In: 2018 International Conference on Content-Based Multimedia Indexing (CBMI). pp.~1--4. IEEE (2018)

\bibitem{han2021dynamic}
Han, Y., Huang, G., Song, S., Yang, L., Wang, H., Wang, Y.: Dynamic neural networks: A survey. IEEE TPAMI  \textbf{44}(11),  7436--7456 (2021)

\bibitem{han2024latency}
Han, Y., Liu, Z., Yuan, Z., Pu, Y., Wang, C., Song, S., Huang, G.: Latency-aware unified dynamic networks for efficient image recognition. IEEE TPAMI  (2024)

\bibitem{he2016deep}
He, K., Zhang, X., Ren, S., Sun, J.: Deep residual learning for image recognition. In: CVPR. pp. 770--778 (2016)

\bibitem{hong2021generic}
Hong, D., Li, C., Wen, L., Wang, X., Zhang, L.: Generic event boundary detection challenge at cvpr 2021 technical report: Cascaded temporal attention network (castanet). arXiv preprint arXiv:2107.00239  (2021)

\bibitem{howard2017mobilenets}
Howard, A.G., Zhu, M., Chen, B., Kalenichenko, D., Wang, W., Weyand, T., Andreetto, M., Adam, H.: Mobilenets: Efficient convolutional neural networks for mobile vision applications. arXiv preprint arXiv:1704.04861  (2017)

\bibitem{hu2018squeeze}
Hu, J., Shen, L., Sun, G.: Squeeze-and-excitation networks. In: CVPR. pp. 7132--7141 (2018)

\bibitem{huang2016connectionist}
Huang, D.A., Fei-Fei, L., Niebles, J.C.: Connectionist temporal modeling for weakly supervised action labeling. In: ECCV. pp. 137--153. Springer (2016)

\bibitem{huang2018multi}
Huang, G., Chen, D., Li, T., Wu, F., van~der Maaten, L., Weinberger, K.: Multi-scale dense networks for resource efficient image classification. In: ICLR (2018)

\bibitem{huynh2023generic}
Huynh, V.T., Yang, H.J., Lee, G.S., Kim, S.H.: Generic event boundary detection in video with pyramid features. arXiv preprint arXiv:2301.04288  (2023)

\bibitem{ioffe2015batch}
Ioffe, S., Szegedy, C.: Batch normalization: Accelerating deep network training by reducing internal covariate shift. In: ICML. pp. 448--456. pmlr (2015)

\bibitem{kang2021winning}
Kang, H., Kim, J., Kim, K., Kim, T., Kim, S.J.: Winning the cvpr'2021 kinetics-gebd challenge: Contrastive learning approach. arXiv preprint arXiv:2106.11549  (2021)

\bibitem{kang2022uboco}
Kang, H., Kim, J., Kim, T., Kim, S.J.: Uboco: Unsupervised boundary contrastive learning for generic event boundary detection. In: CVPR. pp. 20073--20082 (2022)

\bibitem{kingma2014adam}
Kingma, D.P., Ba, J.: Adam: A method for stochastic optimization. arXiv preprint arXiv:1412.6980  (2014)

\bibitem{lea2016segmental}
Lea, C., Reiter, A., Vidal, R., Hager, G.D.: Segmental spatiotemporal cnns for fine-grained action segmentation. In: ECCV. pp. 36--52. Springer (2016)

\bibitem{lei2018temporal}
Lei, P., Todorovic, S.: Temporal deformable residual networks for action segmentation in videos. In: CVPR. pp. 6742--6751 (2018)

\bibitem{li2022structured}
Li, C., Wang, X., Hong, D., Wang, Y., Zhang, L., Luo, T., Wen, L.: Structured context transformer for generic event boundary detection. arXiv preprint arXiv:2206.02985  (2022)

\bibitem{li2022end}
Li, C., Wang, X., Wen, L., Hong, D., Luo, T., Zhang, L.: End-to-end compressed video representation learning for generic event boundary detection. In: CVPR. pp. 13967--13976 (2022)

\bibitem{li2020learning}
Li, Y., Song, L., Chen, Y., Li, Z., Zhang, X., Wang, X., Sun, J.: Learning dynamic routing for semantic segmentation. In: CVPR. pp. 8553--8562 (2020)

\bibitem{lin2019bmn}
Lin, T., Liu, X., Li, X., Ding, E., Wen, S.: Bmn: Boundary-matching network for temporal action proposal generation. In: ICCV. pp. 3889--3898 (2019)

\bibitem{mao2022towards}
Mao, X., Qi, G., Chen, Y., Li, X., Duan, R., Ye, S., He, Y., Xue, H.: Towards robust vision transformer. In: CVPR. pp. 12042--12051 (2022)

\bibitem{ming2021dynamic}
Ming, Q., Zhou, Z., Miao, L., Zhang, H., Li, L.: Dynamic anchor learning for arbitrary-oriented object detection. In: AAAI. vol.~35, pp. 2355--2363 (2021)

\bibitem{nag2023post}
Nag, S., Zhu, X., Song, Y.Z., Xiang, T.: Post-processing temporal action detection. In: CVPR. pp. 18837--18845 (2023)

\bibitem{radvansky2011event}
Radvansky, G.A., Zacks, J.M.: Event perception. Wiley Interdisciplinary Reviews: Cognitive Science  \textbf{2}(6),  608--620 (2011)

\bibitem{shao2020intra}
Shao, D., Zhao, Y., Dai, B., Lin, D.: Intra-and inter-action understanding via temporal action parsing. In: CVPR. pp. 730--739 (2020)

\bibitem{shi2023tridet}
Shi, D., Zhong, Y., Cao, Q., Ma, L., Li, J., Tao, D.: Tridet: Temporal action detection with relative boundary modeling. In: CVPR. pp. 18857--18866 (2023)

\bibitem{shou2021generic}
Shou, M.Z., Lei, S.W., Wang, W., Ghadiyaram, D., Feiszli, M.: Generic event boundary detection: A benchmark for event segmentation. In: ICCV. pp. 8075--8084 (2021)

\bibitem{souvcek2019transnet}
Sou{\v{c}}ek, T., Moravec, J., Loko{\v{c}}, J.: Transnet: A deep network for fast detection of common shot transitions. arXiv preprint arXiv:1906.03363  (2019)

\bibitem{tan2023temporal}
Tan, J., Wang, Y., Wu, G., Wang, L.: Temporal perceiver: A general architecture for arbitrary boundary detection. IEEE TPAMI  (2023)

\bibitem{tang2022progressive}
Tang, J., Liu, Z., Qian, C., Wu, W., Wang, L.: Progressive attention on multi-level dense difference maps for generic event boundary detection. In: CVPR. pp. 3355--3364 (2022)

\bibitem{tang2018fast}
Tang, S., Feng, L., Kuang, Z., Chen, Y., Zhang, W.: Fast video shot transition localization with deep structured models. In: ACCV. pp. 577--592. Springer (2018)

\bibitem{tran2019video}
Tran, D., Wang, H., Torresani, L., Feiszli, M.: Video classification with channel-separated convolutional networks. In: Proceedings of the IEEE/CVF international conference on computer vision. pp. 5552--5561 (2019)

\bibitem{trockman2022patches}
Trockman, A., Kolter, J.Z.: Patches are all you need? arXiv preprint arXiv:2201.09792  (2022)

\bibitem{wang2024towards}
Wang, J., Li, F., An, Y., Zhang, X., Sun, H.: Towards robust lidar-camera fusion in bev space via mutual deformable attention and temporal aggregation. IEEE TCSVT  \textbf{34}(7),  5753--5764 (2024). \doi{10.1109/TCSVT.2024.3366664}

\bibitem{wang2018skipnet}
Wang, X., Yu, F., Dou, Z.Y., Darrell, T., Gonzalez, J.E.: Skipnet: Learning dynamic routing in convolutional networks. In: ECCV. pp. 409--424 (2018)

\bibitem{wang2021adaptive}
Wang, Y., Chen, Z., Jiang, H., Song, S., Han, Y., Huang, G.: Adaptive focus for efficient video recognition. In: ICCV. pp. 16249--16258 (2021)

\bibitem{wang2022adafocus}
Wang, Y., Yue, Y., Lin, Y., Jiang, H., Lai, Z., Kulikov, V., Orlov, N., Shi, H., Huang, G.: Adafocus v2: End-to-end training of spatial dynamic networks for video recognition. In: CVPR. pp. 20030--20040. IEEE (2022)

\bibitem{xie2020spatially}
Xie, Z., Zhang, Z., Zhu, X., Huang, G., Lin, S.: Spatially adaptive inference with stochastic feature sampling and interpolation. In: ECCV. pp. 531--548. Springer (2020)

\bibitem{yang2020resolution}
Yang, L., Han, Y., Chen, X., Song, S., Dai, J., Huang, G.: Resolution adaptive networks for efficient inference. In: CVPR. pp. 2369--2378 (2020)

\bibitem{yang2021condensenet}
Yang, L., Jiang, H., Cai, R., Wang, Y., Song, S., Huang, G., Tian, Q.: Condensenet v2: Sparse feature reactivation for deep networks. In: CVPR. pp. 3569--3578 (2021)

\bibitem{yang2023adadet}
Yang, L., Zheng, Z., Wang, J., Song, S., Huang, G., Li, F.: Adadet: An adaptive object detection system based on early-exit neural networks. IEEE Transactions on Cognitive and Developmental Systems  \textbf{16}(1),  332--345 (2023)

\bibitem{zhang2022actionformer}
Zhang, C.L., Wu, J., Li, Y.: Actionformer: Localizing moments of actions with transformers. In: ECCV. pp. 492--510. Springer (2022)

\bibitem{zheng2023dynamic}
Zheng, Z., Yang, L., Wang, Y., Zhang, M., He, L., Huang, G., Li, F.: Dynamic spatial focus for efficient compressed video action recognition. IEEE TCSVT  (2023)

\end{thebibliography}
\end{document}